# Detection of marine floating plastic using Sentinel-2 imagery and machine learning models

**Srikanta Sannigrahi[a*], Bidroha Basu[a], Arunima Sarkar Basu[a], Francesco Pilla[a]**

[a] School of Architecture Planning and Environmental Policy, University College Dublin, Ireland

Correspondance: srikanta.sannigrahi@ucd.ie

**Abstract**

The increasing level of marine plastic pollution poses severe threats to the marine ecosystem and biodiversity. Open remote sensing data and advanced machine learning (ML) algorithms could be a cost-effective solution for detecting large plastic patches across the scale. The potential application of such resources in detecting and discriminating marine floating plastics are not fully explored. Therefore, the present study attempted to explore the full functionality of open Sentinel satellite data and ML models for detecting and classifying floating plastic debris in Mytilene (Greece), Limassol (Cyprus), Calabria (Italy), and Beirut (Lebanon). Two ML models, i.e. Support Vector Machine (SVM) and Random Forest (RF) were utilized to carry out the classification analysis. In-situ plastic location data was collected from the control experiment conducted in Mytilene, Greece and Limassol, Cyprus, and the same was considered for training the models. The accuracy and performances of the trained models were further tested on unseen new data collected from Calabria, Italy and Beirut, Lebanon.  Both remote sensing bands and spectral indices were used for developing the ML models. A spectral signature profile for plastic was created for discriminating the floating plastic from other marine debris. A newly developed index, kernel Normalized Difference Vegetation Index (kNDVI), was incorporated into the modelling to examine its contribution to model performances. Both SVM and RF were performed well in five models and test case combinations. Among the two ML models, the highest performance was measured for the RF. The inclusion of kNDVI was found effective and increased the model performances, reflected by high balanced accuracy measured for model 2 (~80% to ~98 % for SVM and ~87% to ~97 % for RF). Using the best-performed model, an automated floating plastic detection system was developed and tested in Calabria and Beirut. For both sites, the trained model had detected the floating plastic with ~99% accuracy. Among the six predictors, the FDI was found the most important variable for detecting marine floating plastic. These findings collectively suggest that high-resolution remote sensing imagery and the automated ML models can be an effective alternative for the cost-effective detection of marine floating plastic. Future research will be directed towards collecting quality training data to develop robust automated models and prepare a spectral library for different plastic objects for discriminating plastic from other marine floating debris.

## Introduction

Annually, an estimated 4.4–12.7 million metric tons of plastic are added into the oceans (Borrelle et al., 2017; Jambeck et al., 2015). A 2014 study on marine plastic estimated that nearly 5.25 trillion plastic (weighing 269,000 tons) are floating in the open ocean (Xanthos and Walker, 2017). On average, plastic contributes to 60-80% of the total marine debris, which reach 90-95% in few regions and keep floating in the sea from a hundred to thousands of year due to extended lifespan and durability (Derraik, 2002; Suhrhoff and Scholz-Böttcher, 2016; Walker et al., 2006, 1997; Wang et al., 2016). Like many other global issues such as increases in greenhouse gases, depletion of ozone, large scale accumulation of marine plastic, including the infamous "garbage patches" does not limit within national boundaries (Borrelle et al., 2017). Both microplastic (<5mm) and macro plastic (>5mm) have an insurmountable impact on the marine ecosystem (Borrelle et al., 2017; Green, 2016; Rochman et al., 2016; Sussarellu et al., 2016). Entanglement of marine species by ghost fishnets and ingestion of microplastic by birds and turtles have been commonly reported in the literature (Kühn and van Franeker, 2020; Laist, 1997; Raum-Suryan et al., 2009). The overall economic impact of marine plastic pollution is estimated at ~$13 billion US/year (UNEP, 2014). The accurate detection of marine floating plastic using open-source earth observation data is promising but, at the same time, challenging. Advance machine learning (ML) and deep learning (DL) approaches can be used effectively to detect marine floating plastic automatically and therefore needs detailed investigation to explore the potential cost-effective applications of ML and DL models in tackling marine plastic pollution.

Any types of aquatic and land remote sensing applications often depend on the distinct spectral signatures of earth objects (Biermann et al., 2020a; Topouzelis et al., 2020, 2019). Several spectral bands, i.e. B=490nm, G=560nm, R=665nm, RedEdge2=740nm, NIR=842nm, and SWIR1=1610nm of Sentinel 2A/B L1C top-of-atmosphere (TOA) data have been used for detecting marine floating plastic in various studies (Topouzelis et al., 2020; Biermann et al., 2020; Themistocleous et al., 2020). Moreover, several previous research has suggested the usability and functional capability of optical/infrared bands in detecting marine floating debris, including floating plastics (Acuña-Ruz et al., 2018; Biermann et al., 2020a; Maximenko et al., 2019, 2012; Pichel et al., 2012, 2007; Topouzelis et al., 2020, 2019). The spectral reflectance of clear and white plastic objects is found high in the NIR channel and often reflects flat signals in the visible spectrum (Themistocleous et al., 2020; Topouzelis et al., 2020, 2019). In comparison, the clear water absorbs light in the NIR region and found highly reflective in the Blue region. However, the type of plastics often determines the remote sensing reflectance in different spectrum. For instance, Topouzelis (2019, 2020) study has shown that blue plastic bags' reflectance was high in the blue and NIR channels, and the same was found low in the green and red channels. The plastic bottles with clear white colour emit flat signal throughout the visible spectrum and highly reflective in the NIR spectrum (Dierssen and Garaba, 2020; Garaba and Dierssen, 2020; Goddijn-Murphy et al., 2018; Goddijn-Murphy and Dufaur, 2018; Masoumi et al., 2012).

Apart from the uses of discrete spectral bands, different spectral indices, i.e. Floating Debris Index (FDI) (Biermann et al., 2020a), Normalized Difference Vegetation Index (NDVI) (Carlson and Ripley, 1997; Tucker et al., 2005), Plastic Index (PI) (Themistocleous et al., 2020), etc. have also been utilized for detecting marine floating debris. Additionally, previous research also noted that choosing the correct atmospheric correction (AC) processor is arguably the most crucial part of image pre-processing, and it often appears to be challenging to extract water surface information from at-sensor or top-of-atmosphere radiance (Biermann et al., 2020b; Martins et al., 2017; Vanhellemont and Ruddick, 2021, 2018; Wang et al., 2019; M. A. Warren et al., 2019). The retrieval accuracy and correct representation of land/water surface from the at-sensor radiance signals strongly depend on the accuracy of atmospheric correction algorithms (Bi et al., 2018; Ilori et al., 2019; Katkovsky et al., 2018).

The potential application of advanced ML and artificial intelligence techniques to detect and discriminate marine floating plastics is not fully explored yet. Therefore, the present study was aimed to detect the marine floating plastic in Mytilene (Greece), Limassol (Cyprus), Calabria (Italy), and Beirut (Lebanon) with the help of advanced ML models and open access in-situ plastic and satellite data. A Python-based automated floating plastic detection system was developed for predicting floating marine plastic effectively. The performances of the newly developed classifier were tested in two test sites (Calabria and Beirut) to ensure the reliability of the classifier. The present study has also explored the importance of nonlinear remote sensing indices in detecting floating marine plastic. A newly developed nonlinear vegetation index, called kernel Normalized Difference Vegetation Index, was incorporated into the ML modelling to examine its contribution to overall model performances.

**2. Materials and method**

**2.1 Selection of sites**

The present research has utilized the in-situ plastic locations data originated from the control experiments conducted by Topouzelis et al. (2020, 2019) in Mytilene (Greece) and Themistocleous et al. (2020) in Limassol (Cyprus) on various dates in 2018 and 2019 (**Fig. 1**). These controlled experiments were mainly conducted to complement the satellite-based plastic detection process and used as reference data to develop a spectral profile of plastic objects. Limassol port, Cyprus, was considered in this study because of the availability of validated and geocoded plastic locations data originated from the controlled experiment conducted by Themistocleous et al. (2020) on 15 December 2018. This study (Themistocleous et al.) conducted a detailed experiment to detect floating marine plastic using open-source Sentinel 2A/B data. A 3 m * 10 m plastic litter target consists of water bottles with size 0.5 litre and 1.5-litre was deployed in Limassol coast, Cyprus. Topouzelis et al. (2020, 2019) and the teams from the University of Agene, Greece, have taken the novel and first of its kind initiative to detect marine floating plastics using open Satellite remote sensing data. This control experiment was conducted by deploying plastic targets made of bottles, bags, and fishing nets at Tsamakia beach, Mytilene, Greece, on different dates in 2018 and 2019.

For testing the accuracy and performances of the ML models, another two sites, Calabria (Italy) and Beirut (Lebanon), were used in this study (**Fig. 2**). In October 2018, heavy rainfall events experienced in the Calabria region in Southern Italy. This caused severe flooding across the downstream areas and added a substantial amount of trash into the ocean. Several Sentinel 2A/B scenes that covered the region were collected and processed with atmospheric correction processors to test the trained classifier's performance. There are ample evidences of marine plastic pollution in Beirut, and therefore the same was also considered in the present study. Relevant information was collected from different social media platforms that support the existence of marine litter in Beirut.

## 2.2 Sentinel 2 L1C data collection and pre-processing

The present study has utilized Sentinel 2A/B 13 bands Multispectral imagery (MSI) level 1C data collected from Copernicus Open Access Hub (https://scihub.copernicus.eu/). A total of 8 Sentinel 2A/B scenes (5 for Mytilene, 1 for Limassol, 1 for Calabria, and 1 for Beirut) were retrieved from Copernicus Open Access Hub and processed with European Space Agency provided Sentinel Application Platform (SNAP) v. 8 software. The basic image pre-processing such as spatial subset, resampling was done in SNAP software. Among the 13 MSI bands, only 4 bands, i.e. Band 4 (Red), Band 6 (Red Edge 2), Band 8 (NIR), and Band 11 (Short Wave Infrared 1), were utilized in this study for developing the spectral signature of plastic and performing multiparameter supervised models. The land and cloud mask was applied to masked out the land area from the analysis. The details of the Sentinel 2 bands, spatial resolution, spectral resolution, temporal resolution, and image acquisition information is provided in **Table. S1**.

Since the primary aim of the present study was to retrieve the remote sensing reflectance (Rrs) of clean water, selecting the right atmospheric correction processors was of utmost importance for this research. Sentinel 2A/B, Level 1C top-of-the atmosphere (TOA) reflectance data, was converted to surface reflectance estimates using ACOLITE and Case 2 Regional Coast Color (C2RCC) atmospheric correction processors. The ESA's default Sen2Cor atmospheric correction processor has not been used in this study as the same was found ineffective and performed poorly over the open water surface (Martins et al., 2017; Vanhellemont and Ruddick, 2021, 2018; Wang et al., 2019; M. A. Warren et al., 2019). Additionally, earlier studies (Themistocleous et al., 2020; Topouzelis et al., 2020, 2019) have found that Sen2Cor generated high reflectance values for the targeted plastic object in places where the controlled experiment was conducted (Mytilene, Greece) and therefore could introduce uncertainty in the detection analysis. The C2RCC AC processor developed by (Doerffer and Schiller, 2007) is a Neural Network (NN) based AC processor used many water-specific application across the scale (Bi et al., 2018; Pereira-Sandoval et al., 2019; Renosh et al., 2020). C2RCC originally developed based on a large set of NN led inverted radiative transfer simulation (Renosh et al., 2020).

## 2.3 Preparation of training and validation dataset

A total of 54 validated plastic pixels collected from Mytilene (Greece) and Limassol (Cyprus) were used to train the supervised ML models. For conducting the classification in Mytilene and Limassol, out of the 54 data, 70 % (38) was used for training, and 30% (16) was used for validating the results. Additionally, for classifying the individual Sentinel scene retrieved for Mytilene and Limassol, a total of 12 (for 7 June 2018), 8 (for 18 April 2019), 8 (for 3 May 2019), 12 (for 18 May 2019), 8 (for 28 May 2019), and 6 (for 12 December 2018) samples were used to label the plastic class. For water class, a total of 68 (for 7 June 2018), 72 (for 18 April 2019), 72 (for 3 May 2019), 67 (for 18 May 2019), 62 (for 28 May 2019), and 43 (for 12 December 2018) samples were used for classification.

To test the model performances and assess the potential application of the developed model on real-world data, all the 54 in-situ geocoded plastic data was used to train the ML models. To handle the class imbalance problem, an equal number of water samples (58) were collected from Mytilene and Limassol, and the same was used to label the water feature. The model that performed most accurately on ground data (collected from Mytilene and Limassol) was considered for predicting floating plastics in the testing sites (Beirut and Calabria). A total of 42 and 33 samples were collected from Calabria and Beirut for labelling the plastic feature for modelling. For the water class, a total of 82 and 53 samples were collected. The details of training and testing data used in the ML models are provided in **Fig. 1** and **Fig. 2**.

FDI and band reflectance in NIR and Red Edge 2 was used to discriminate the plastic pixels from Sentinel true colour (RGB) image. A true-colour image was prepared for all four locations for the mentioned dates. A visual inspection was carried out to detect the target plastic pixels and verify the difference between the plastic pixels' spectral responses and the surrounding water pixels. Followed by, the spectral responses of plastic pixels in the NIR and Red spectral region were checked to analyse the deviation of the spectral signal emitted from plastics. Using the geocoded (Latitude and Longitude) information of the targeted plastic pixels available from Topouzelis et al. (2020) for Mytilene, Greece and Themistocleous et al. (2020) for Limassol, Cyprus, the surface reflectance values of the 4 spectral bands, i.e. Red, Red

Edge2, NIR, and SWIR1, and four remote sensing indices, i.e. FDI, NDVI, Plastic Index (PI), and Kernel NDVI (kNDVI) were prepared.

**2.4 Remote sensing spectral indices**

Among the remote sensing spectral indices, FDI, PI, NDVI, and kernel NDVI were used in this study as input parameter for developing the supervised ML models. These indices were chosen based on their superior performances in the detection of floating marine plastics in Greece (Topouzelis et al., 2019, 2020), Cyprus (Themistocleous et al., 2020), and Durban (Biermann et al., 2020). Among these indices, kNDVI has not been applied before for detecting marine floating plastic. Therefore, to our best knowledge, this is the first study exploring the usability of nonlinear NDVI (kNDVI) to detect floating plastic in the open ocean. FDI was calculated by measuring the difference between NIR and baseline reflectance of NIR as follows:

$$FDI = R_{rs,}NIR - R'_{rs,}NIR$$
$$R'_{rs,}NIR = R_{rs,}RE2 + (R_{rs,}SWIR1 - R_{rs,}RE2) * \frac{(833-665)}{(1610-665)} * 10 \qquad (1)$$

Where $FDI$ is the floating debris index, $R_{rs,}NIR$, $R_{rs,}RE2$, $R_{rs,}SWIR1$ are the remote sensing reflectance of NIR, Red Edge 2, and SWIR 1 bands, $R'_{rs,}NIR$ is the baseline reflectance of NIR, 665, 833, and 1610 are the central wavelength (nm) of Red, NIR, and SWIR 1 bands, respectively.

The plastic index (Themistocleous et al. 2020) has also been utilized in this research for modelling and prediction of plastic locations. The PI was developed based on the remote sensing reflectance in two spectral regions, i.e. Red and NIR. The PI can be calculated as follows:

$$PI = \frac{R_{rs,}NIR}{(R_{rs,}NIR + R_{rs,}RED)} \tag{2}$$

Where NIR and Red refer to the pixel's reflectance in NIR and Red spectrum.

Following, two vegetation indices, such as NDVI and the newly proposed nonlinear vegetation index, i.e. Kernel NDVI, have also been measured using the pixel reflectance in NIR and Red spectral region. NDVI can be calculated as follows:

$$NDVI = \frac{R_{rs,}NIR - R_{rs,}RED}{R_{rs,}NIR + R_{rs,}RED} \tag{3}$$

Where $R_{rs,}NIR - R_{rs,}RED$ refers to the remote sensing reflectance of NIR and Red bands.

To eliminate the problem of linear assumption in NDVI calculation, a nonlinear kernel approach was adopted to map the spectral bands to a high-dimensional feature map. Additionally, to convert the linear NDVI to a nonlinear one, NDVI was mapped in Hilbert spaces, and kernelization was done using the radial basis function (RBF) (Camps-Valls et al., 2021). The kernel NDVI (kNDVI) was calculated as follows:

$$kNDVI = \frac{k(n,n) - k(R_{rs,}NIR, R_{rs,}RED)}{k(n,n) + k(R_{rs,}NIR, R_{rs,}RED)} \tag{4}$$

Where $n, R_{rs,}NIR$ and $n, R_{rs,}RED$ refers to the remote sensing reflectance in NIR and Red channels, $k$ is the kernel function that measures the similarity between the two bands, i.e. NIR and Red in the case of NDVI. The kernel function $k$ was calculated using the RBF kernel as follows:

$$k(a,b) = \exp(-(a-b)^2 / ((2\sigma^2))) \tag{5}$$

Where $a, b$ are the two bands in the case of NDVI, and $\sigma$ parameter determines the distance between NIR and Red bands. The kernelization was further simplified as:

$$kNDVI = \frac{1-k(R_{rs,}NIR, R_{rs,}RED)}{1+k(R_{rs,}NIR, R_{rs,}RED)} = \tanh\left(\left(\frac{R_{rs,}NIR - R_{rs,}RED}{2\sigma}\right)^2\right) \quad (6)$$

The fixing of $\sigma$ parameter was done as follows:

$\sigma = 0.15(R_{rs,}NIR + R_{rs,}RED)$ and this $\sigma$ parameter fixing further simplified the kernelization as follows:

$$kNDVI = \tanh(NDVI^2) \quad (7)$$

**2.5 ML modelling for detection marine floating plastics**

Since the present study has carried out the supervised classification for multiple dates and locations, it has appeared to be safe to discard the linear assumption in the modelling and to adopt non-linear models to obtain better precision in modelling. The present study has adopted a few most used nonlinear supervised ML models, i.e. Support Vector Machine (SVM) and Random Forest (RF), for detecting marine floating plastics and classifying the regions into two classes, plastic and clean water. Four remote sensing spectral indices, i.e. FDI, NDVI, PI, kNDVI, and three spectral bands, i.e. Red Edge 2, NIR, and SWIR1, were used as input parameters for the ML modelling. The other bands and spectral indices were not utilized in the modelling as the same were found not relevant for classification and subsequent analysis. Random forests (RFs) are an example of ensemble modelling, which denotes that RF produces multiple numbers of estimators during the modelling and finally combine them to generate the final prediction (Breiman, 2001). In RF modelling, the dataset splits into training and testing data. The training dataset is further separated as bootstrap samples utilised in each decision tree, and out-of-bag samples for evaluating the performance of RF models The SVM is a widely used supervised ML algorithm applied to classify objects and predict variable using a standard (non)linear regression approach (Vapnik, 2000). The performance of the SVM model is often determined by the selection of appropriate kernel function, i.e. linear, polynomial, sigmoid, and radial basis functions (RBF). Among the four kernel functions, RBF was used in the present study considering its superior performance in both classification and regression.

Hyperparameter tuning was performed to optimize the parameters used in ML models and ensure that the models do not overfit the results. Hyperparameters such as *ntree* (number of trees), *mtry* (number of splits) in RF, *C* parameter and *gamma* parameter in SVM and weights in ANN are the settings that can modulate the model performances, and proper attention, therefore, needs to pay to adjust these parameters setting for obtaining unbiased model outcomes and better accuracy in modelling. An experimental design was developed, consisting of a combination of ML algorithms and training/testing data to optimize the hyperparameters. Among the two main hyperparameter tuning methods, i.e. random search and grid search, the present study has utilized the grid search function considering its better performances in many relevant application. All the possible combinations of hyperparameters are tested in the given ranges for each parameter in the grid search process. Finally, the ML models were trained using the best hyperparameter combinations. The hyperparameter tuning was done using the *caret* R package, and *sci-kit learn* package in Python. The details of the optimized hyperparameters used in the modelling are given in **Table. 1** and **Table. 2**.

## 2.6 Experimental design

The entire modelling and subsequent analysis were carried out using a combination of parameters setting and test cases. **Fig. 3** shows the successive steps and approaches that have followed in carrying out the entire analysis. Both pixel and non-pixel based classifications were performed to assess the overall performances of the models and to examine the models' efficiency in detecting plastic pixels from Sentinel 2A/B 10m scene without producing many false-positive estimates. A total of five Sentinel scenes for Mytilene, Greece and one Sentinel scene for Limassol, Cyprus, have been used for carrying out the image-based classification. The water pixels used for training/testing the models have also been verified with the spectral signature profile of water developed by Bierman et al. (2020). The spatial location of the training/testing data can be seen in **Fig. 1, Fig. 2**. A total of 25 parameter/testing combinations were prepared for conducting the non-pixel level classification. The details of these model combinations are provided in **Table. S2, S3**.

An experimental setup has been developed to assess the model performances in different training and testing dataset combinations, consisting of five test case scenarios with varied training and testing data for conducting the classification using in-situ data in Mytilene and Limassol. For test case 1, an equal number of samples, i.e. 54 for plastic and 54 for water, were used to develop the ML models. For the remaining test cases, the samples for water class were increased up to two times (for test case 2), three times (for test case 3), four times (for test case 4), and five times 270 (for test case 5), for analyzing the model's sensitivity to different training/testing combinations. The number of plastic data remains the same for all test cases as we have only 54 in-situ data for plastic. In all these test case combinations, 70% of data were used for training the models, and the remaining 30% of data was used to validate the models. Both RF and SVM models were performed in five parameters and five test case combinations, thus collectively generating a matrix consisting of 25 model classification output.

After comparing the model performances in 25 model/test case combinations, the best performing model with the highest accuracy score on in-situ data was chosen to assess its performance and application on real-world data. The domain and parameter approximation of the best performing model was kept remain the same while executing its performances on real-world data.

**2.7 Model performance and accuracy assessment**

Model performances and classification accuracy was assessed using different accuracy matrices, i.e. Kappa, Mcnemar test, sensitivity, specificity, precision, recall, F1, and balanced accuracy. The details of these accuracy matrices can be found in (Ebrahimy et al., 2021; Grabska et al., 2020; Vogeler et al., 2016). The ML classification was performed in R programming software using *caret* for hyperparameter tuning, *e1071* and *kernlab* for SVM, *randomForest* for RF, *nnet* and *NeuralNetTools* for ANN, and *readxl, reshape2, ggplot2, rgdal, raster, sp, grid, gridextra, RStoolbox, dplyr, rattle* for other uses. *Sci-kit learn* Python package was used for testing the trained model in real-world data. Kernel NDVI was computed using *raster* R package. All the statistical analysis and model accuracy assessment was done in R software. Remote sensing data was handled using ArcGIS Pro v2.7 and Sentinel Application Platform (SNAP) v.8. ACOLITE and C2RCC atmospheric correction was done in ACOLITE v. 20210114.0 software and SNAP v.8. The density scatter plot was prepared using Python v3.8.

## 3 Results

### *3.1 Spectral signature of plastics and atmospheric correction of satellite data*

The floating marine debris patches were extracted from Sentinel 2 images for Calabria and Beirut are presented in **Fig. 4** and **Fig. 5**. Histogram stretching was applied to the input data to enhance the contrast stretch of the raster. For both Calabria and Beirut, the histogram stretched enhanced data was found to be highly effective to detect floating debris. The spectral signature profile for different plastic objects, i.e. plastic bags, plastic bottles, and fishnets, is presented in **Fig. 6**. All the 54 plastic pixels used in preparing the spectral signature of plastic were segregated into five categories, 0-10%, 10-20%, 20-30%, 30-40%, and >40%, respectively. These percentage values represent the proportion of plastic coverage in a Sentinel pixel. In **Fig. 6**, the left panel shows the reflectance properties of the five plastic categories in different channels, i.e. Blue (B2), Green (B3), Red (B4), Red Edge 1 (B5), Red Edge 2 (B6), Red Edge 3 (B7), NIR (B8), Red Edge 4 (B8A), SWIR 1 (B11), and SWIR 2 (B12), respectively. In comparison, the plot in the right panel shows the average reflectance of each plastic categories in a varied spectral region. Both plastic bags and plastic bottles exhibited high reflectance in the NIR region, while it has shown a low to flat reflectance in the Red Edge region. In **Fig. 6c**, the reflectance values of the seven confirmed pixels (plastic percentage ~>28%) are presented. Out of the seven pixels, 5 are composed of plastic bottles, and the remaining two pixels were composed of plastic bag and fishnet. The seven confirmed pixels averaged have shown distinct reflective properties and showed a peak reflectance in the NIR spectrum. These specific reflectance properties of plastic could help to discriminate plastic pixels from other floating marine debris. However, the intensity of reflectance in the NIR region depends on the proportion of plastic present in a pixel.

*3.2 Performances of ML models in detecting marine floating plastics*

The results of the image-based classification are presented in **Fig. 7**. Model 1 to 5 represents the varied parameters setups that were used to develop the models. On the 7 June 2018 date's classification, both RF and SVM perform most accurately in model 4 and model 5 combinations. While the same performed poorly and produces high false-positive estimates in model 3 combination (**Fig. 7**). On 18 April 2019 date classification, the SVM model has shown the best performances in model 4 and model 5 combinations, while it has incorrectly classified a few water pixels as plastic in model 1, model 2, and model 3 combinations. On the other hand, RF performs accurately in model 2, 3, and 5 combinations and performs moderately in model 1 combination. Additionally, the RF model performs poorly in model 3 combination, and falsely classified a number of water pixels as plastic that eventually reduced the classification accuracy (**Fig. S1**). For 3 May 2019 classification, the SVM model performs moderately in model 2, 4 and 5 combinations (**Fig. S1**). However, it has produced erroneous classification estimates in model 1 and 3 combinations. While the RF model accurately classified the plastics and performed well in all model combinations except (model 1 and model 3) (**Fig. S1**). For 15 December 2018 classification, both SVM and RF produced high classification accuracy and successfully detected the floating plastics in all model combinations except model 5 (**Fig. 7**). The usability of FDI in developing ML models and discriminating floating plastic is evaluated and presented in **Fig. 8**. FDI was utilized on Mytilene and Limassol on multiple dates, and for all dates, FDI accurately detected the validated plastic pixels (**Fig. 8**). The FDI based manual detection of floating plastic was also applied in Beirut and Calabria, and the same was used to cross-check the ML detected floating plastics.

The accuracy matrices, including kappa, Mcnemar P-Value, sensitivity, specificity, precision, recall, F1, balanced accuracy, etc., were measure for the ML models with different model and test case combinations (**Fig. 9, Table 3**). For model 1 combination, the kappa and balanced accuracy score was found highest for RF. Additionally, among the five test case combinations, the kappa and balanced accuracy of the SVM model was measured highest in test case 2 combination, while the same is found highest in test case 4 combinations for the RF model. Overall, the RF model has produced the highest accurate classification outputs and perform most accurately in model 1 (**Table 3**). For model 2, the SVM and RF models have produced high accuracy and accurately classified the plastic and water class for all test case combinations. SVM was performed well in test case 3 combination and performed relatively poorly in test case 4 and test 5 combinations. On the other hand, the RF model correctly predicted the plastic and water class in test case 2 and test case 3. Among the three ML models, the RF model's accuracy was highest in the model 2 combination (**Table 3**). For model 3, SVM performed with high precision in test case 2, while the same performed poorly in test case 3 combinations. However, for RF, the classification accuracy was found different from other ML models. The highest kappa and balanced accuracy of the RF model is measured in test case 5, followed by test case 3, test case 2, test case 4, and test case 1, respectively (**Table 3**). Both the models were not performed well in model 4 and model 5 combinations and produced poor accuracy score for all five test case combinations (**Table 3**).

### 3.3. *Performance evaluation of ML models on real-world data*

The accuracy and performances of the ML model were assessed, and results are provided in **Fig. 10** and **Table. 4**. Among the two ML models, i.e. SVM and RF, RF performed most accurately on the in-situ data and produced the highest accuracy scores for most of the model/test case combinations. The trained ML model classified all the 42 and 82 plastic and water pixels accurately and obtained 100% accuracy for Calabria. For Beirut, among the 33 plastic pixels, 31 were correctly classified by the trained model. While, for the same location, the model correctly predicted all the water pixels with 100 accuracy (**Fig. 10**). The variability of the OOB error rate with the changes in number of trees is presented in **Fig. 10**. For both Calabria and Beirut, the OOB error rate started diminishing after ~40 – 50 estimators. The variable importance score, shown through mean decrease accuracy, indicated that FDI is the most important variable for detecting floating plastic using an automated ML model (**Fig. 10**). Band 8, PI, and kNDVI were also found to be important and, therefore, could be effective to detect floating plastic from Sentinel 2 data.

## 4 Discussion

The present study observed a noticeable difference between the spectral responses of plastic bags, bottles, and fishnets. Reflections of plastic bottles in the NIR region are much higher than those of plastic bags and fishnets. The thickness of plastic material and degree of weathering often determines the spectral reflectance shape of plastic. The earlier study (Moshtaghi et al., 2021) observed that absorption in 1192–1215, 1660 nm and 1730 nm channel is most suitable for discriminating polymer object from the mixed element. Among them, the absorption feature in 1730 nm would be most effective for differentiating plastic from wood (Moshtaghi et al., 2021).

The colour of plastic objects is also an important factor as it would be hard to discriminate coloured plastic signals from water in the visible spectrum. Additionally, brown and blue coloured plastics signals are often mixed with turbid plume and clean water, making it hard to discriminate them from open ocean water (Moshtaghi et al., 2021). Garaba and Dierssen (2020) study on hyperspectral analysis on marine harvested, washed-ashore, and virgin plastics noted that a high absorption in NIR-SWIR spectrum centred ranges from 931nm to 2313nm. Additionally, they also noted that a decrease in reflectance magnitude once the dry plastic gets dampened. Since we did not have any in-situ reflectance data for plastics, we relied on the atmospherically corrected remote sensing reflectance estimates and utilized the same for both developing the spectral signature of plastic and constructing the ML models.

Both the ML models utilized in this study, i.e. SVM and RF, were performed accurately and classified the plastic pixels with satisfactory accuracy. SVM model's performances were improved significantly with increasing sample size for a few model/test case combinations. Additionally, it has been noted that for the SVM model, remote sensing spectral indices alone can not produce a high model performance. The RF model performed most accurately and classified the plastic pixels with less errors and uncertainty in all model and test case combinations. For a few cases, the performance of the RF model found decreased when sample sizes were increased. As discussed above, the design of test cases with imbalanced plastic and water data might have been the reason for this. The spectral indices alone were found insignificant in improving RF model performance conducted on in-situ data, as seen in several model/test case combinations. This suggests that both remote sensing bands and spectral indices are crucial for developing an automated classification system for detecting marine floating plastic. Considering the model performances on real-world data, the RF model performed accurately on real-world data collected from Calabria and Beirut. An average 99% accuracy was achieved for the trained classifier performed in Calabria and Beirut on different dates.

In previous studies, attempts have been made to employ the open Satellite data in finding ocean plastic across the scale (Biermann et al. 2020, Basu et al., 2021). Since most of the previous efforts have relied on remote sensing reflectance of satellite data, chances of getting erroneous or mixed signals are high if proper attention on pre-processing, including atmospheric and geometric correction, is not provided. Using the coupled application of remote sensing bands and indices, Biermann et al. (2020) successfully detected floating plastic in Accra (Ghana), Gulf Island (Canada), Scotland (UK), and Da Nang (Vietnam). A total of six marine debris, i.e. plastic, sea spume, timber, sargassum seagrass, water, pumice, etc., were categorized and classified using Sentinel data based on the spectral signature of floating debris. Using this spectral analysis approach, the present study has developed an automated supervised machine learning classifier that detected the floating plastic with 99% accuracy on real-world data. In our study, very high model performances were achieved for both the testing sites that mainly happened due to the less number of classes (plastic and water) were considered in the assessment. Basu et al. (2021) study on developing clustering algorithms for detecting floating plastic also obtained ~97% accuracy in detecting floating plastic using validated plastic data and noted that open Sentinel 2 10m data could be effectively used for discriminating floating marine debris. However, this study (Basu et al.) also emphasized on collecting more quality training data for developing any plastic detection model.

Developing a data-driven spectral signature library for different marine debris is outmost important in remote sensing-based floating plastic detection as in real-world scenario, floating plastic is often mixed with other debris such as seagrass, algae, timber, pumice, etc., and this complex aggregation of debris patches often makes the detection process challenging. Since the present study had taken a comprehensive approach to carefully verify the spectral signature of the suspected plastic pixels used for both training and testing the models, the chances of getting erroneous false positive alarms in modelling were less prevalent. However, as a previous study (Biermann et al. 2020) mentioned that floating plastic and sea plume have almost similar reflectance properties in VNIR and SWIR regions, more training data needs to be collected to reduce the false alarms in the future floating plastic detection analysis. Though the present study has adopted two most used supervised ML models, i.e. SVM and RF, a thorough assessment by incorporating a number of different genre models need to be done to better understand the strength and weakness of different algorithms in detecting floating plastic. The potential application of different advanced deep learning algorithms and architectures, such as DeepLab, RetinaNet, UNET, YOLO, etc. in detecting marine floating plastic, needs to be explored to develop a robust data-driven plastic detection system that eventually helps the concerned authorities and decision-makers to adopt relevant strategies to address the marine litter problem.

**Conclusion**

The present study utilized high-resolution Sentinel 2A/B satellite imagery and advanced machine learning models to detect and classify marine floating plastics in Mytilene (Greece), Limassol (Cyprus), Calabria (Italy) and Beirut (Lebanon). Distinct and noticeable differences between the spectral responses of plastic bags, bottles, and fishnets are observed in the present research. The reflectance of plastic bottles in the NIR channel is found much higher than those of plastic bags and fishnets. The thickness of plastic material, degree of weathering, and bio-fouling could explain these differences. It needs to be noted that the age and color of plastics can also be a crucial factor in developing ML models as the virgin and weathered floating plastic can reflect lights differently. A total of 54 validated in-situ plastic locations data was collected for Mytilene and Limassol on various dates in 2018 and 2019 for training and developing the models. Since classification accuracy of any ML models largely controlled by both quality and quantity of training data, the present study has adopted a fivefold model/test case combinations to assess the sensitivity of the model performances to different modelling set-ups. The SVM model's performance in different model and test case combinations suggests that the remote sensing spectral indices alone can not produce a high model performance for the SVM model. RF model performance in varied model/test case combinations indicates that both remote sensing bands and spectral indices equally important for detecting floating plastic. Among the two ML models, RF performed most accurately for nearly all model and test case combinations and therefore applied to detect floating plastic on real-world data collected from Calabria and Beirut. Using the trained model, we were able to discriminate floating plastic with 99% accuracy. Among the six key predictors used in this study, FDI is the most important variable and with the highest variable importance score. This suggests that future modelling can be done by considering only a few parameters that significantly reduce the time and cost needed to prepare model data. Also, the developed model can be applied to any location with similar seawater characteristics and external perturbing factors.

**Acknowledgements**

This publication has emanated from research conducted with the financial support of Science Foundation Ireland under Grant number 20/FIP/PL/8752. The authors are grateful to Konstantinos Topouzelis, Kyriacos Themistocleous, and Lauren Biermann for providing the - in-situ plastic data. We would like to thank the European Space Agency (ESA) for providing the Sentinel 2A/B images.

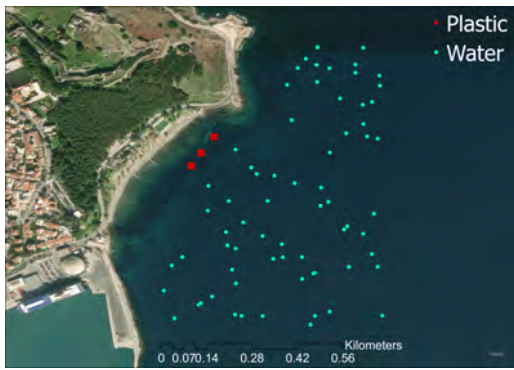 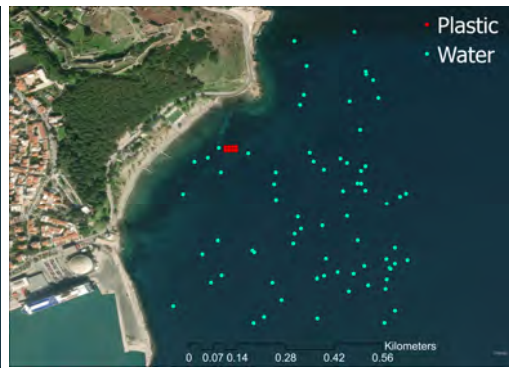 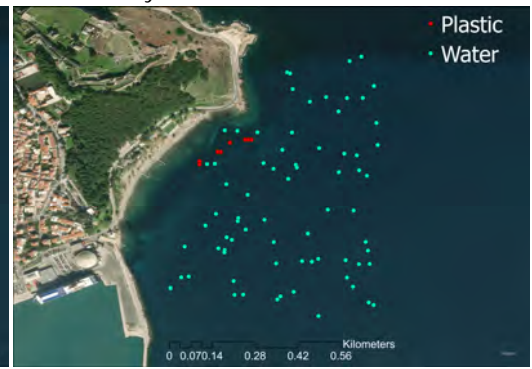
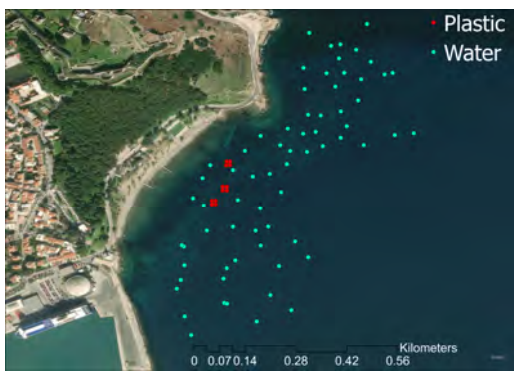 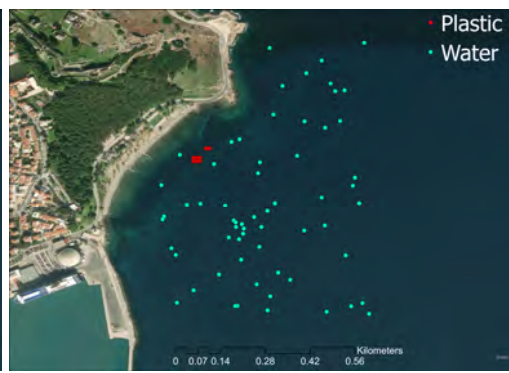 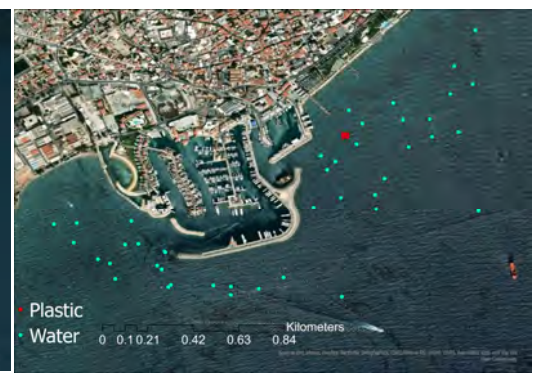

**Fig. 1** Spatial distribution of the in-situ validated plastic data used for training the supervised machine learning models, i.e. support vector machine and random forest. The water training samples are represented by cyan color, and training samples used for plastic are presented in red color.

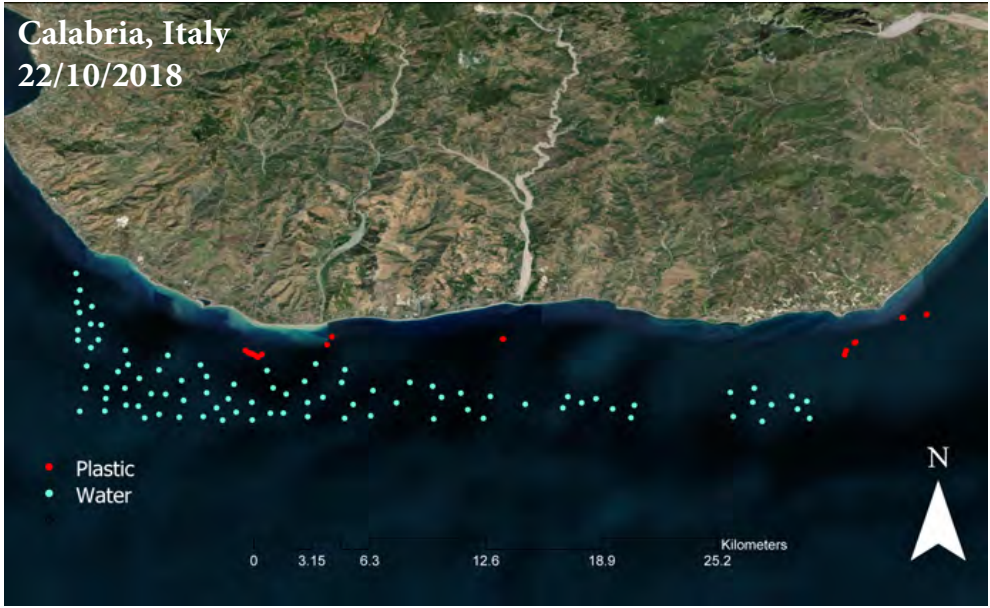
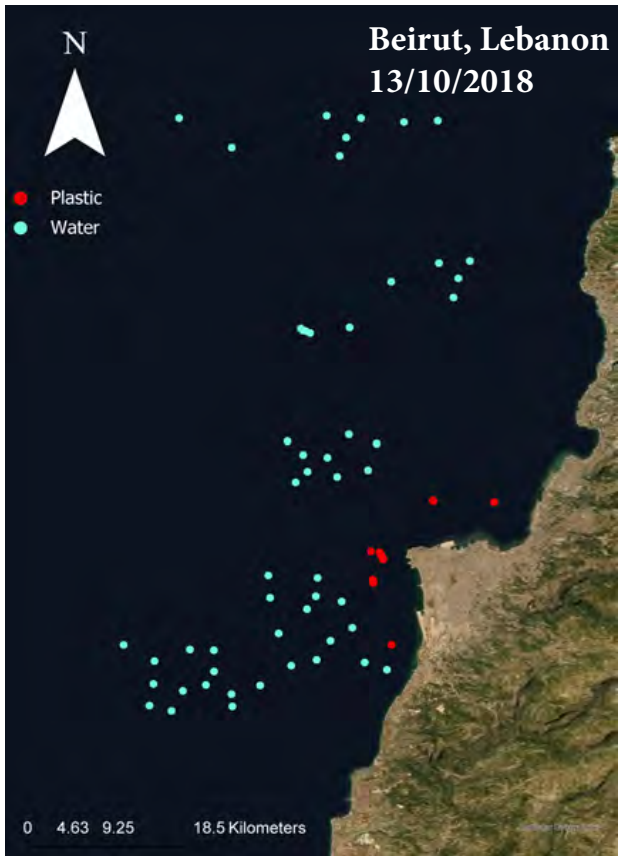

**Fig. 2** Spatial locations of the testing data used for testing the supervised model performances. The water and plastic training samples are represented by cyan and red color.

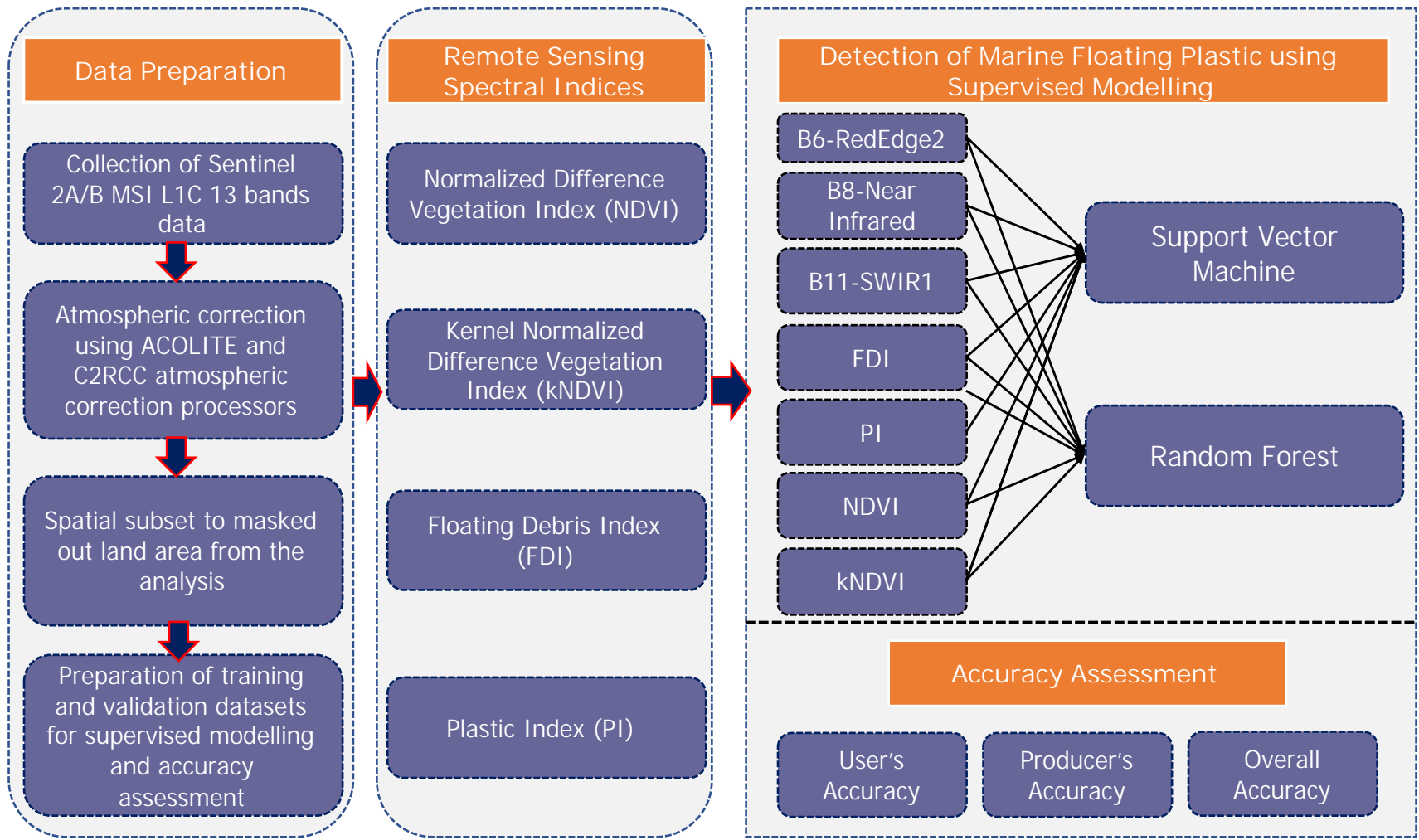

**Fig. 3** Methodological flow charts showing the successive steps adopted for carrying out the marine floating detection analysis.

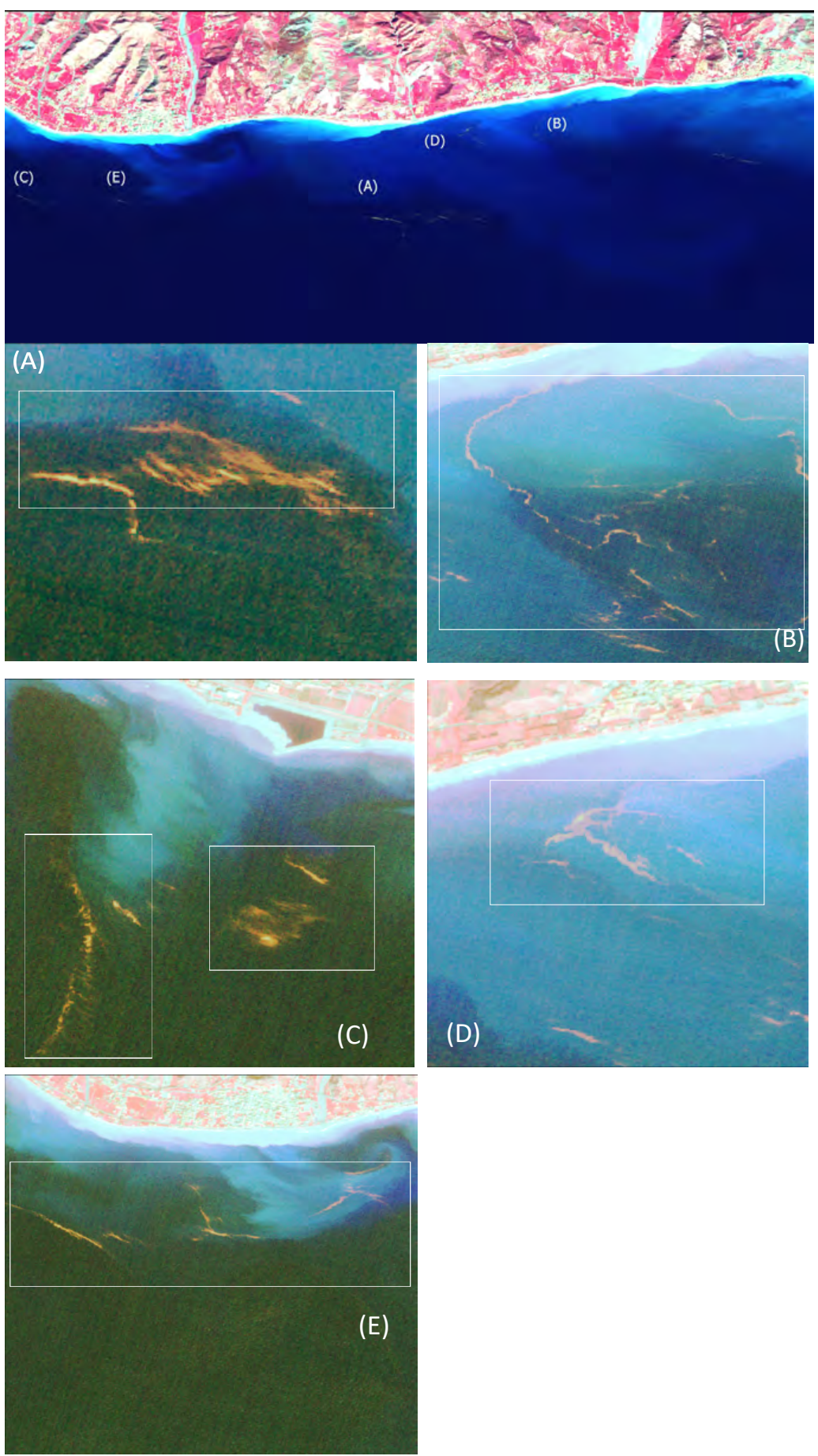

Fig. 4 Location of floating marine debris detected from Sentinel 2 image on 22 October, 2018 in Calabria, Italy. Panel A, B, C, D, and E refers to the exact location of the marine litter found within 10km of the Calabria coast, Italy.

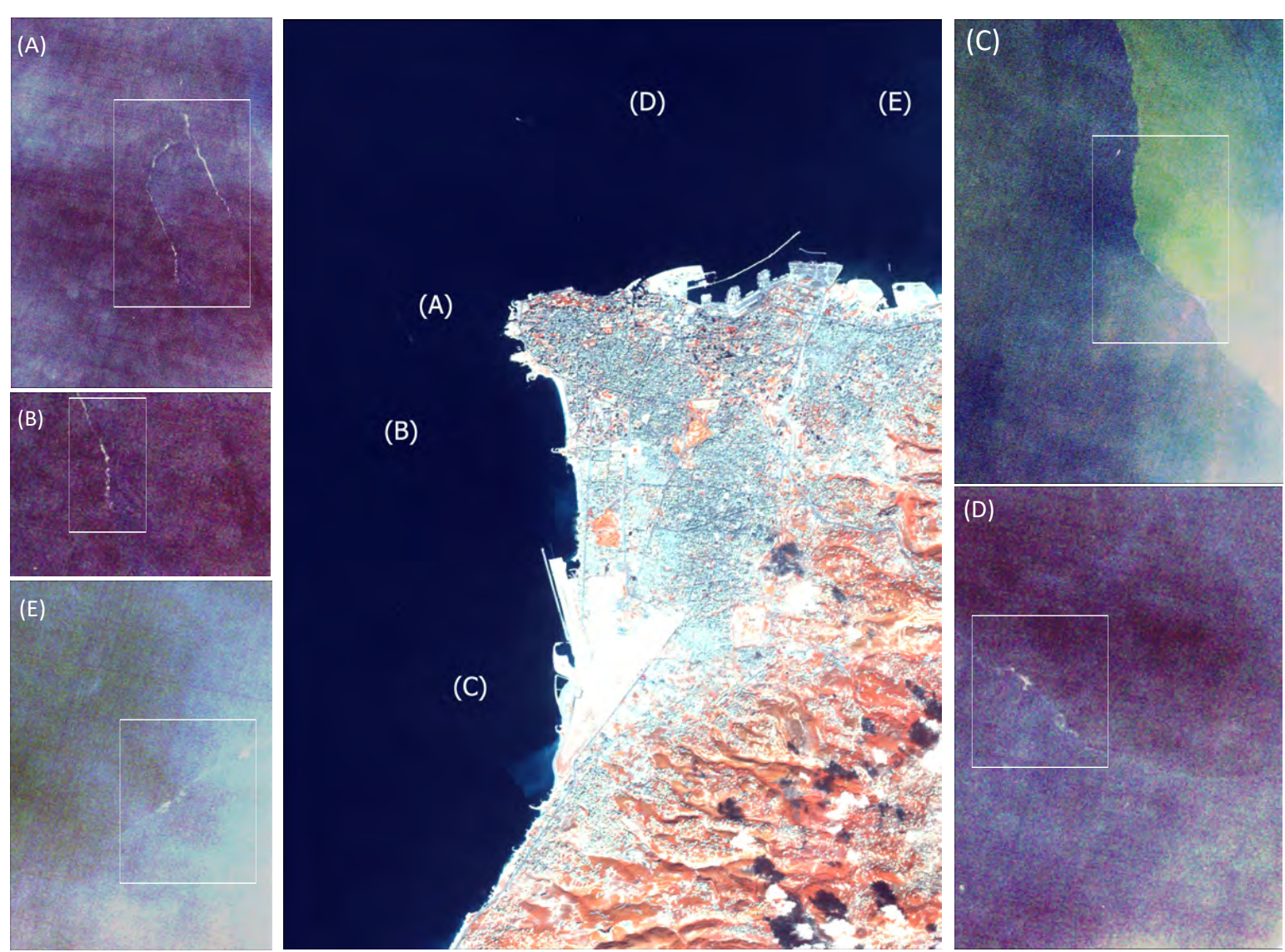

**Fig. 5** Location of floating marine debris detected from Sentinel 2 image on 13 October, 2018 in Beirut, Lebanon. Panel A, B, C, D, and E refers to the exact location of the marine litter floating within 10km of the Beirut coast, Lebanon.

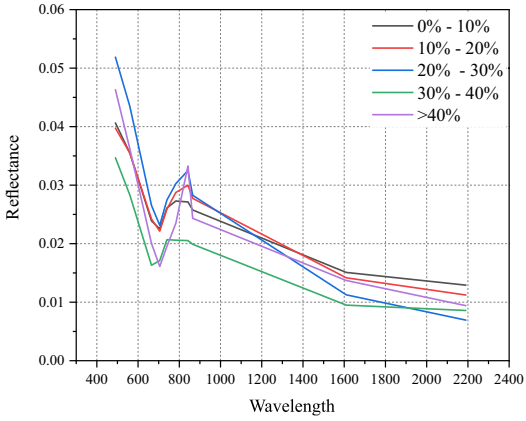
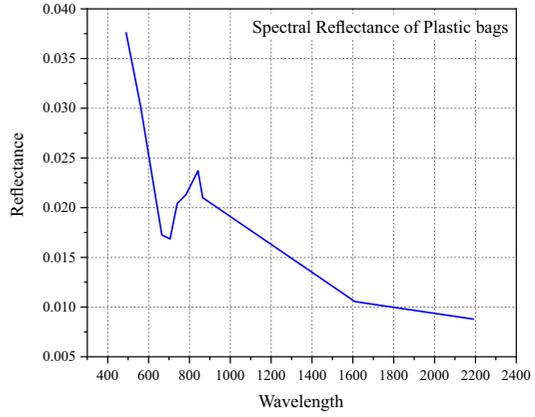

**(a) Plastic bags**

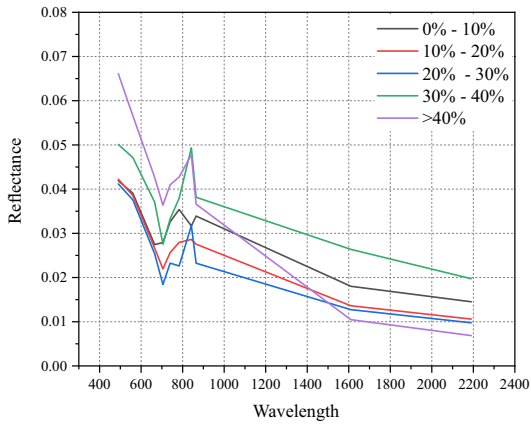
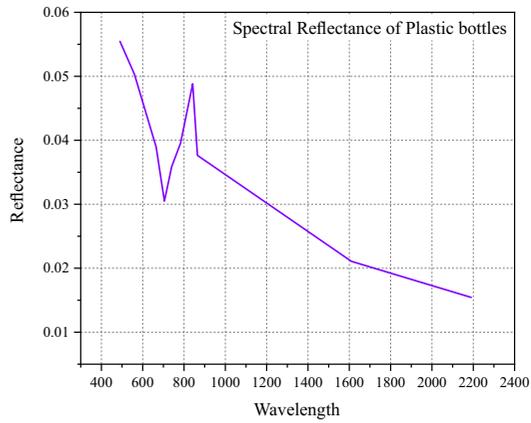

**(b) Plastic bottles**

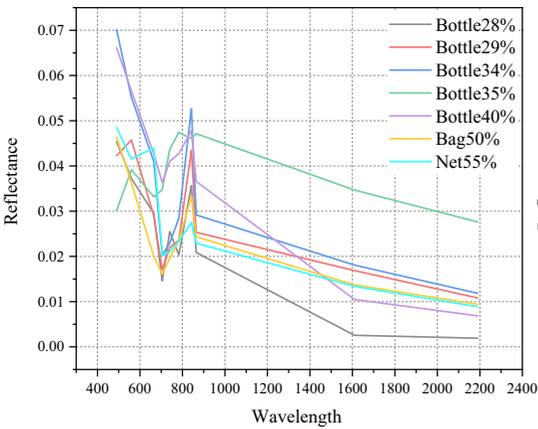
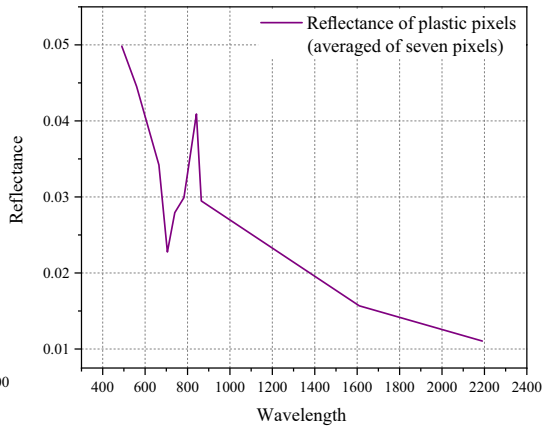

**(c) Confirmed plastic pixels**

**Fig. 6** Spectral signature profile of plastic targets made of plastic bottles, plastic bags, fishnets, etc. All the plastic pixels were grouped into five categories, plastic concentration 0 – 10%, 10% - 20%, 20% - 30%, 30% – 40%, and >40%, respectively (**Fig. 6a, b**). Finally, the spectral signature of the confirmed plastic pixels (the plastic percentage is >25% for bottles, >50% for bags and nets) were calculated (**Fig. 6c**).

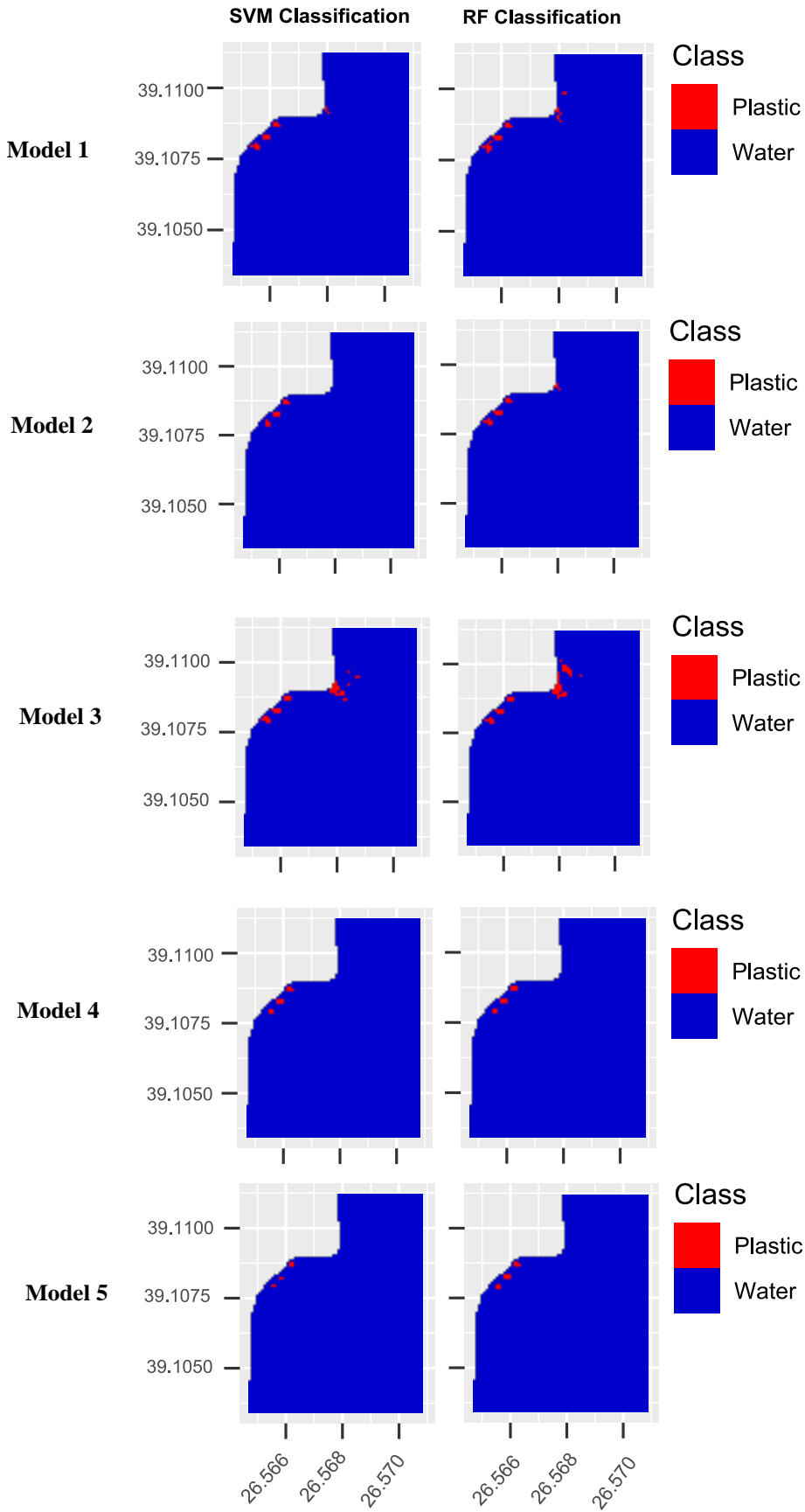

**Fig. 7 Continued........**

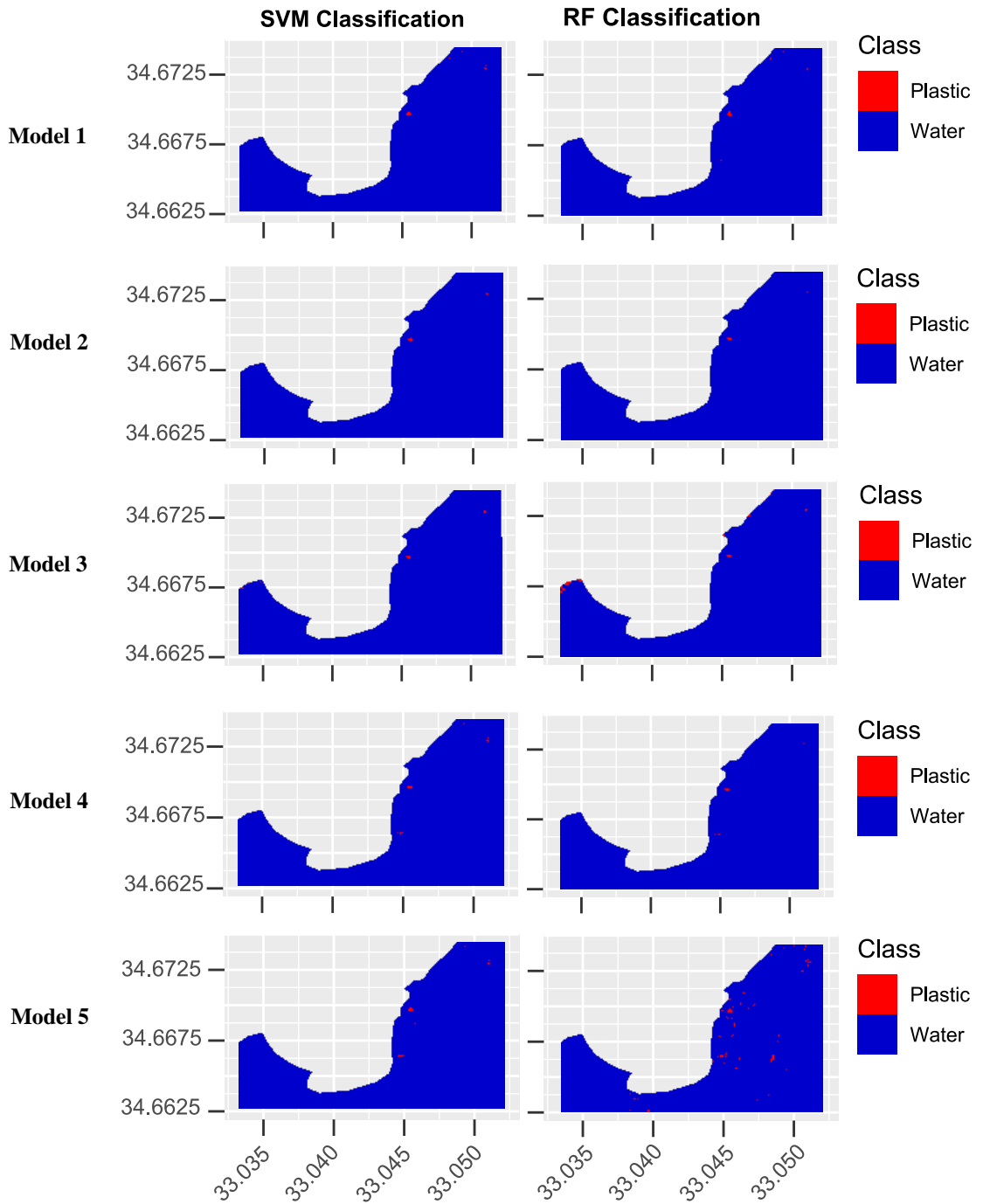

Fig. 7 Shows the results of machine learning supervised models, SVM and RF, performed in Mytilene, Greece and Limassol, Cyprus on 07/06/2018 and 15/12/2018 and the spatial location of the plastics shown in red colour). Blue color represent the pixels classified as water, and red color indicates the pixel classified as plastic by the models. All the three models were performed in five alternative combinations, such as Model 1 constructed using B6, B8, B11, FDI, PI, and NDVI), Model 2 (constructed using B6, B8, B11, FDI, PI, and kernel NDVI), Model 3 (constructed using B6, B8, B11, FDI), Model 4 (constructed using FDI, PI, and NDVI), and Model 5 (constructed using FDI, PI, and kernel NDVI).

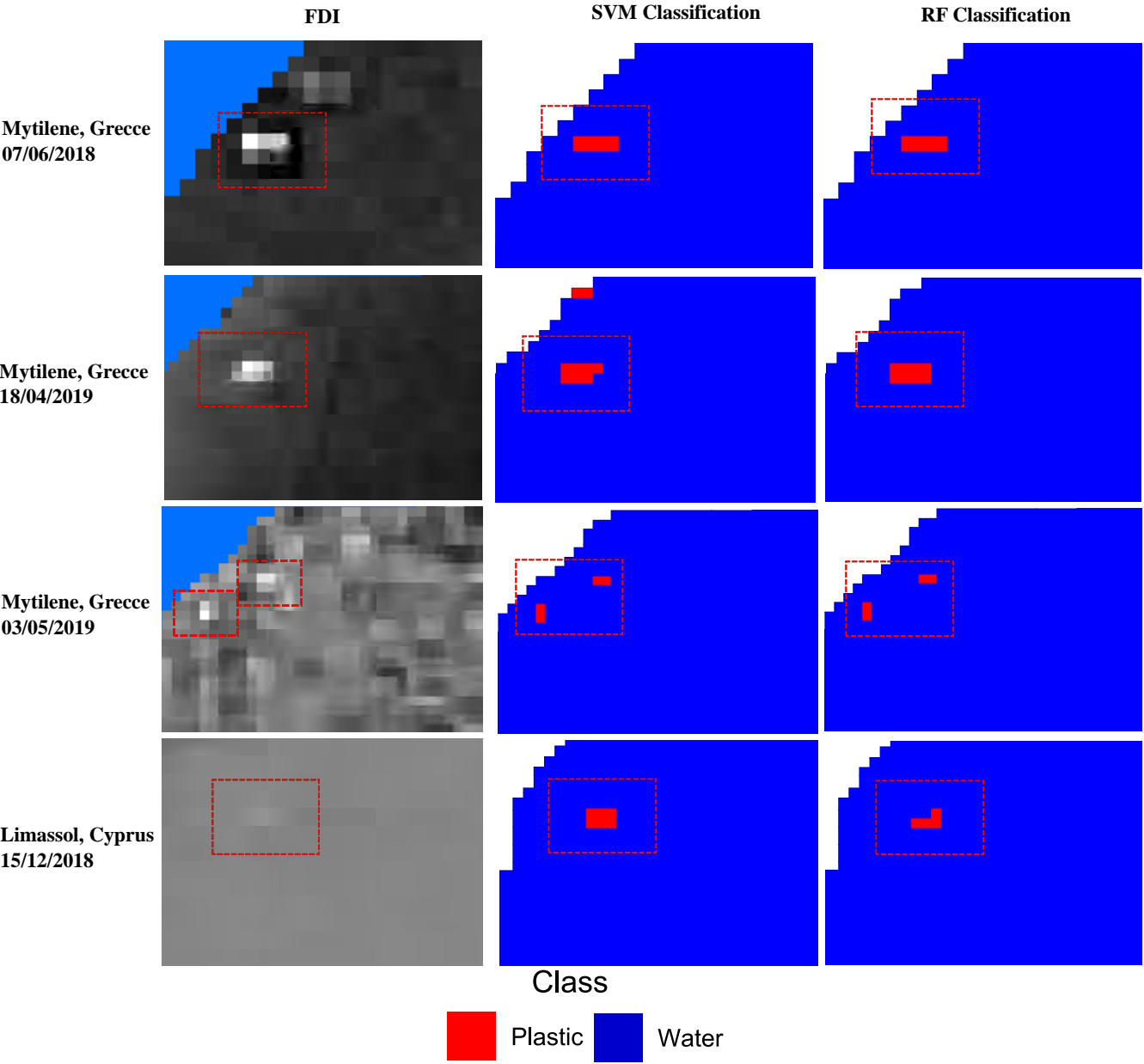

Fig. 8 Floating Debris Index and predicted plastic pixels using SVM and RF model on different dates in Mytilene, Greece, and Limassol, Cyprus.

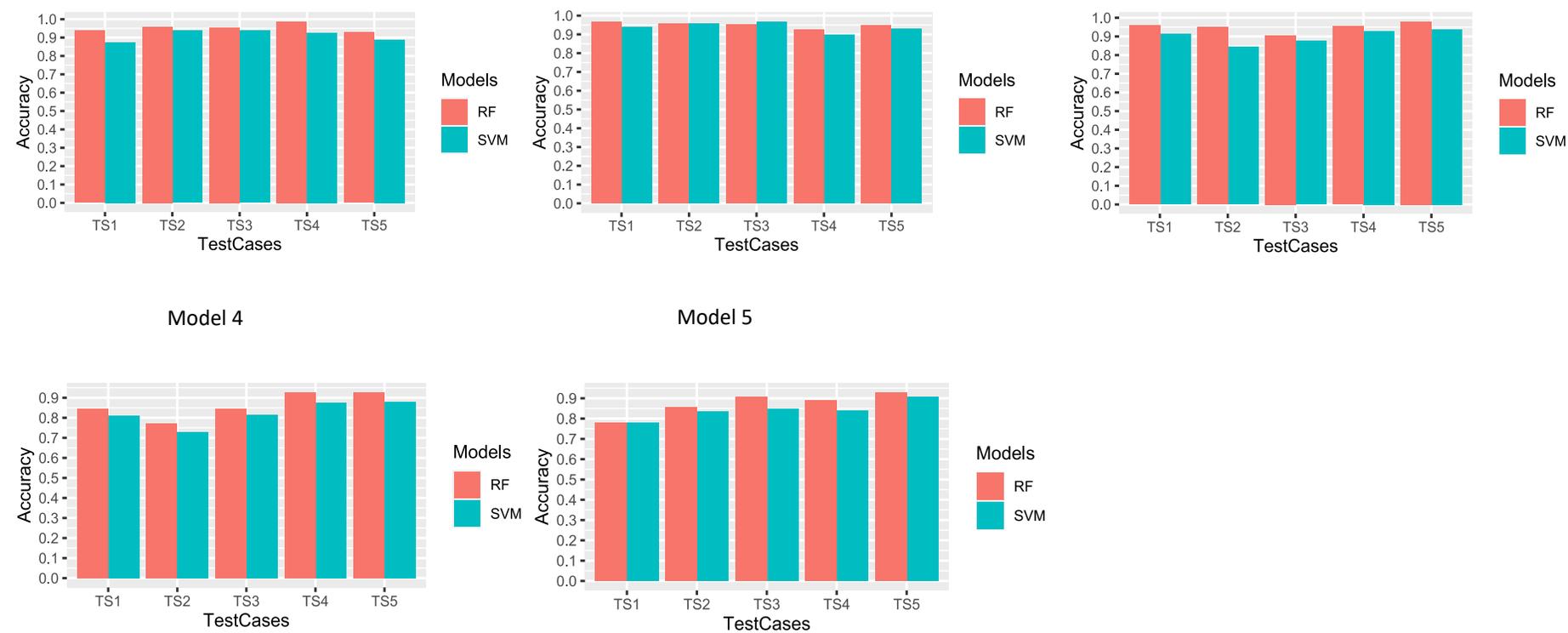

Fig. 9 Classification accuracy of the three machine learning models, SVM – Support Vector Machine, RF – Random Forest, performed in five test case combinations. TS 1-5 refers to different test cases utilized for validating the models. Model 1-5 was constructed using varied parameter combinations, such as – Model 1 = Red Edge 2, NIR, SWIR, FDI, PI, and NDVI; Model 2 = Red Edge 2, NIR, SWIR, FDI, PI, and kernel NDVI; Model 3 = Red Edge 2, NIR, SWIR, FDI; Model 4 = FDI, PI, and NDVI; Model 5 = FDI, PI, and kernel NDVI, respectively. TS 1-5 refers to different test cases utilized for validating the models. NIR – Near Infra Red, SWIR – Shortwave Infrared, FDI – Floating Debris Index, PI – Plastic Index, NDVI – Normalized Difference Vegetation Index, kNDVI – Kernel NDVI.

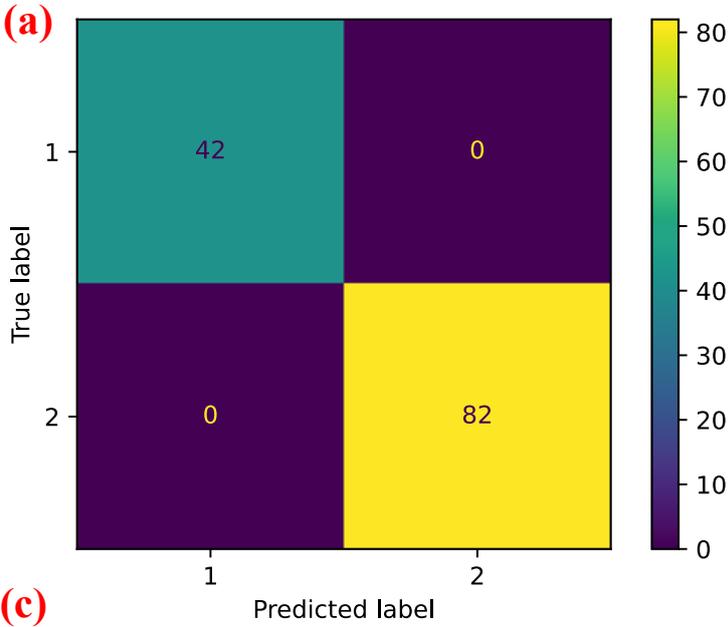
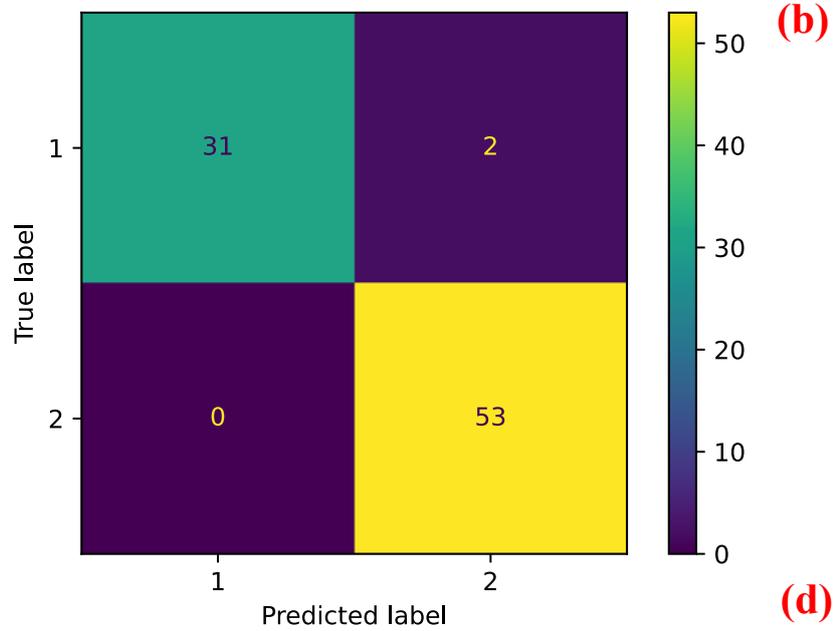
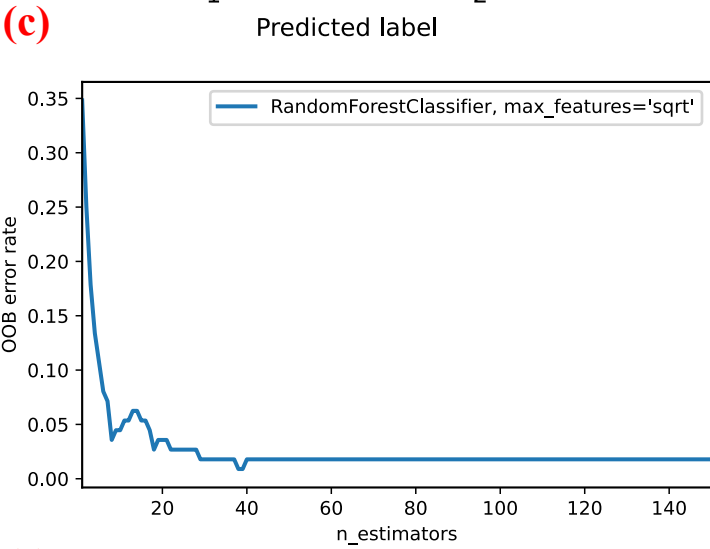
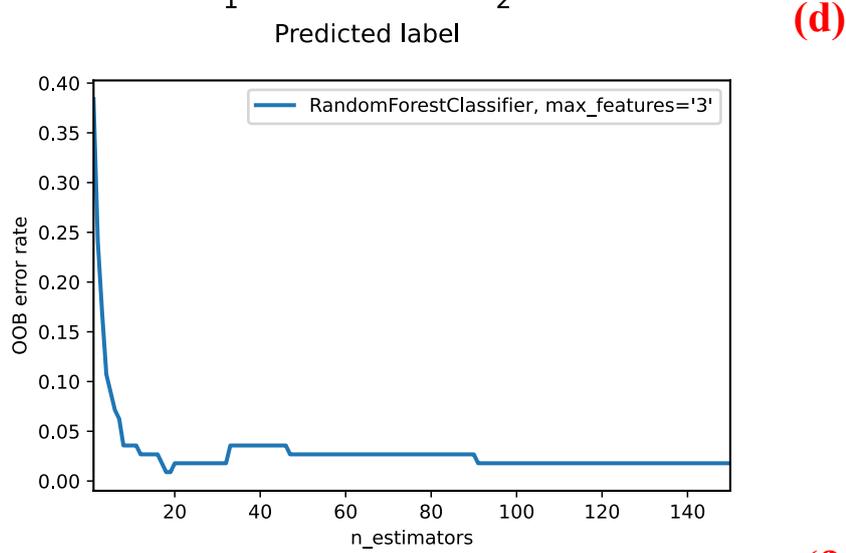
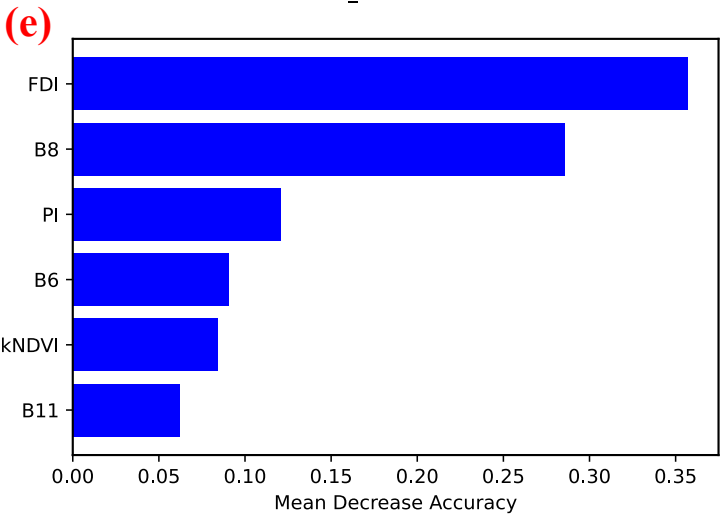
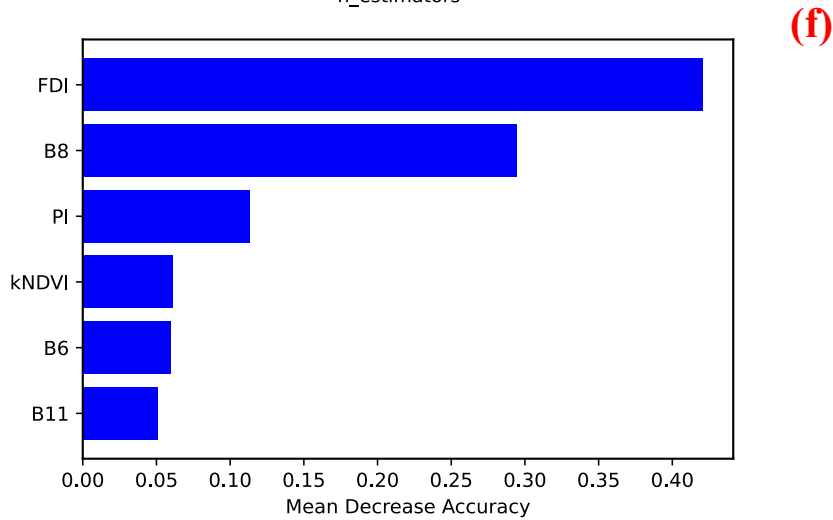

**Fig. 10** Confusion matrix, OOB error score, and variable importance analysis summarizing the performances of RF classification model. The diagonal elements of panel A and B refers to the observations correctly classified as true level. The off diagonal values are reflecting the values misclassified by the classifier. Panel C and D refers to the OOB error rate and its variability in different ntrees parametrization. Panel C and D refers to the variable importance score of the explanatory variables used in the supervised modelling. **1 = Plastic, 2 = Water**

**Table 1** Descriptions of machine learning model hyperparameters tuning at different modelling setups. Model 1-5 was constructed using varied parameter combinations, such as – Model 1 = Red Edge 2, NIR, SWIR, FDI, PI, and NDVI; Model 2 = Red Edge 2, NIR, SWIR, FDI, PI, and kernel NDVI; Model 3 = Red Edge 2, NIR, SWIR, FDI; Model 4 = FDI, PI, and NDVI; Model 5 = FDI, PI, and kernel NDVI, respectively. TS 1-5 refers to different test cases utilised for validating the models. NIR – Near Infra Red, SWIR – Shortwave Infrared, FDI – Floating Debris Index, PI – Plastic Index, NDVI – Normalized Difference Vegetation Index, kNDVI – Kernel NDVI, TS – Test Case, SVM – Support Vector Machine, RF – Random Forest, ntrees – Number of Tress, mtry – Number of Split at each node, ML – Machine Learning, C – Cost factor, SV – Support Vectors.

| Models | Test cases | ML Algorithms | |
|---|---|---|---|
| | | SVM | RF |
| Model 1 | TS1 | Sigma - 0.09, C - 10, SV - 37 | ntrees - 500, mtry - 1 |
| | TS2 | Sigma - 0.09, C - 10, SV - 39 | ntrees - 500, mtry - 3 |
| | TS3 | Sigma - 0.09, C - 10, SV - 48 | ntrees - 500, mtry - 1 |
| | TS4 | Sigma - 0.07, C - 10, SV - 57 | ntrees - 500, mtry - 2 |
| | TS5 | Sigma - 0.09, C - 10, SV - 52 | ntrees - 500, mtry - 4 |
| Model 2 | TS1 | Sigma - 0.09, C - 10, SV - 36 | ntrees - 500, mtry - 1 |
| | TS2 | Sigma - 0.09, C - 10, SV - 40 | ntrees - 500, mtry - 1 |
| | TS3 | Sigma - 0.09, C - 10, SV - 56 | ntrees - 500, mtry - 3 |
| | TS4 | Sigma - 0.09, C - 10, SV - 50 | ntrees - 500, mtry - 3 |
| | TS5 | Sigma - 0.09, C - 10, SV - 56 | ntrees - 500, mtry - 3 |
| Model 3 | TS1 | Sigma - 0.09, C - 10, SV - 39 | ntrees - 500, mtry - 4 |
| | TS2 | Sigma - 0.09, C - 10, SV - 47 | ntrees - 500, mtry - 1 |
| | TS3 | Sigma - 0.09, C - 10, SV - 54 | ntrees - 500, mtry - 2 |
| | TS4 | Sigma - 0.09, C - 10, SV - 58 | ntrees - 500, mtry - 1 |
| | TS5 | Sigma - 0.09, C - 10, SV - 58 | ntrees - 500, mtry - 2 |
| Model 4 | TS1 | Sigma - 0.09, C - 10, SV - 50 | ntrees - 500, mtry - 8 |
| | TS2 | Sigma - 0.09, C - 8, SV - 49 | ntrees - 500, mtry - 1 |
| | TS3 | Sigma - 0.03, C - 2, SV - 73 | ntrees - 500, mtry - 1 |
| | TS4 | Sigma - 0.09, C - 10, SV - 65 | ntrees - 500, mtry - 1 |
| | TS5 | Sigma - 0.09, C - 6, SV - 70 | ntrees - 500, mtry - 1 |
| Model 5 | TS1 | Sigma - 0.09, C - 8, SV - 46 | ntrees - 500, mtry - 1 |
| | TS2 | Sigma - 0.09, C - 10, SV - 51 | ntrees - 500, mtry - 12 |
| | TS3 | Sigma - 0.09, C - 8, SV - 61 | ntrees - 500, mtry - 10 |
| | TS4 | Sigma - 0.09, C - 10, SV - 56 | ntrees - 500, mtry - 2 |
| | TS5 | Sigma - 0.09, C - 10, SV - 66 | ntrees - 500, mtry - 1 |

**Table. 2** Description of the optimized hyperparameters used for developing the final model.

| Hyperparameters | Value |
|---|---|
| N Estimators | 100 |
| Bootstrap | TRUE |
| Criterion | gini |
| Max_Features | sqrt |
| Max_Depth | 6 |
| Verbose | 1 |
| Min_Samples_Leaf | 2 |
| Max_Leaf_Nodes | 8 |
| OOB_score | 1 |
| Min_Samples_Split | 2 |
| Random_State | 0 |

**Table 3** Accuracy estimates of supervised machine learning models, i.e. Support Vector Machine (SVM), Random Forest (RF). Test cases 1-5 refers to different test cases utilised for validating the models. Model 1-5 was constructed using varied parameter combinations, such as, Model 1 = Red Edge 2, NIR, SWIR, FDI, PI, and NDVI; Model 2 = Red Edge 2, NIR, SWIR, FDI, PI, and kernel NDVI; Model 3 = Red Edge 2, NIR, SWIR, FDI; Model 4 = FDI, PI, and NDVI; Model 5 = FDI, PI, and kernel NDVI, respectively.

| Models | Accuracy Estimates | Test case1 SVM | Test case1 RF | Test case2 SVM | Test case2 RF | Test case3 SVM | Test case3 RF | Test case4 SVM | Test case4 RF | Test case5 SVM | Test case5 RF |
|---|---|---|---|---|---|---|---|---|---|---|---|
| **Model 1** | Accuracy | 0.875 | 0.938 | 0.938 | 0.958 | 0.938 | 0.954 | 0.925 | 0.988 | 0.888 | 0.929 |
| | Kappa | 0.750 | 0.875 | 0.866 | 0.909 | 0.827 | 0.873 | 0.727 | 0.960 | 0.530 | 0.757 |
| | Mcnemar P Value | 0.617 | 1.000 | 0.248 | 0.480 | 0.617 | 1.000 | 0.041 | 1.000 | 0.228 | 0.450 |
| | Sensitivity | 0.938 | 0.938 | 1.000 | 1.000 | 0.813 | 0.875 | 0.625 | 0.938 | 0.500 | 0.875 |
| | Specificity | 0.813 | 0.938 | 0.906 | 0.938 | 0.980 | 0.980 | 1.000 | 1.000 | 0.963 | 0.939 |
| | Precision | 0.833 | 0.938 | 0.842 | 0.889 | 0.929 | 0.933 | 1.000 | 1.000 | 0.727 | 0.737 |
| | Recall | 0.938 | 0.938 | 1.000 | 1.000 | 0.813 | 0.875 | 0.625 | 0.938 | 0.500 | 0.875 |
| | F1 | 0.882 | 0.938 | 0.914 | 0.941 | 0.867 | 0.903 | 0.769 | 0.968 | 0.593 | 0.800 |
| | Balanced Accuracy | 0.875 | 0.938 | 0.953 | 0.969 | 0.896 | 0.927 | 0.813 | 0.969 | 0.732 | 0.907 |
| **Model 2** | Accuracy | 0.938 | 0.969 | 0.958 | 0.958 | 0.969 | 0.954 | 0.900 | 0.925 | 0.929 | 0.949 |
| | Kappa | 0.875 | 0.938 | 0.909 | 0.909 | 0.920 | 0.878 | 0.655 | 0.766 | 0.732 | 0.798 |
| | Mcnemar P Value | 1.000 | 1.000 | 0.480 | 0.480 | 0.480 | 1.000 | 0.289 | 1.000 | 1.000 | 0.371 |
| | Sensitivity | 0.938 | 1.000 | 1.000 | 1.000 | 1.000 | 0.938 | 0.625 | 0.813 | 0.750 | 0.750 |
| | Specificity | 0.938 | 0.938 | 0.938 | 0.938 | 0.959 | 0.959 | 0.969 | 0.953 | 0.963 | 0.988 |
| | Precision | 0.938 | 0.941 | 0.889 | 0.889 | 0.889 | 0.882 | 0.833 | 0.813 | 0.800 | 0.923 |
| | Recall | 0.938 | 1.000 | 1.000 | 1.000 | 1.000 | 0.938 | 0.625 | 0.813 | 0.750 | 0.750 |
| | F1 | 0.938 | 0.970 | 0.941 | 0.941 | 0.941 | 0.909 | 0.714 | 0.813 | 0.774 | 0.828 |
| | Balanced Accuracy | 0.938 | 0.969 | 0.969 | 0.969 | 0.980 | 0.948 | 0.797 | 0.883 | 0.857 | 0.869 |
| **Model 3** | Accuracy | 0.844 | 0.906 | 0.938 | 0.958 | 0.877 | 0.954 | 0.913 | 0.950 | 0.929 | 0.980 |
| | Kappa | 0.688 | 0.813 | 0.862 | 0.903 | 0.620 | 0.883 | 0.673 | 0.828 | 0.683 | 0.925 |
| | Mcnemar P Value | 1.000 | 1.000 | 1.000 | 0.480 | 0.077 | 0.248 | 0.023 | 0.134 | 0.023 | 1.000 |
| | Sensitivity | 0.875 | 0.938 | 0.938 | 0.875 | 0.563 | 1.000 | 0.563 | 0.750 | 0.563 | 0.938 |
| | Specificity | 0.813 | 0.875 | 0.938 | 1.000 | 0.980 | 0.939 | 1.000 | 1.000 | 1.000 | 0.988 |
| | Precision | 0.824 | 0.882 | 0.882 | 1.000 | 0.900 | 0.842 | 1.000 | 1.000 | 1.000 | 0.938 |
| | Recall | 0.875 | 0.938 | 0.938 | 0.875 | 0.563 | 1.000 | 0.563 | 0.750 | 0.563 | 0.938 |
| | F1 | 0.848 | 0.909 | 0.909 | 0.933 | 0.692 | 0.914 | 0.720 | 0.857 | 0.720 | 0.938 |
| | Balanced Accuracy | 0.844 | 0.906 | 0.938 | 0.938 | 0.771 | 0.969 | 0.781 | 0.875 | 0.781 | 0.963 |
| Model 4 | Accuracy | 0.813 | 0.844 | 0.729 | 0.771 | 0.815 | 0.846 | 0.875 | 0.925 | 0.878 | 0.929 |
| | Kappa | 0.625 | 0.688 | 0.291 | 0.421 | 0.370 | 0.602 | 0.545 | 0.754 | 0.439 | 0.732 |
| | Mcnemar P Value | 1.000 | 0.074 | 0.027 | 0.070 | 0.009 | 0.752 | 0.114 | 0.683 | 0.043 | 1.000 |
| | Sensitivity | 0.813 | 1.000 | 0.313 | 0.438 | 0.313 | 0.750 | 0.500 | 0.750 | 0.375 | 0.750 |
| | Specificity | 0.813 | 0.688 | 0.938 | 0.938 | 0.980 | 0.878 | 0.969 | 0.969 | 0.976 | 0.963 |
| | Precision | 0.813 | 0.762 | 0.714 | 0.778 | 0.833 | 0.667 | 0.800 | 0.857 | 0.750 | 0.800 |
| | Recall | 0.813 | 1.000 | 0.313 | 0.438 | 0.313 | 0.750 | 0.500 | 0.750 | 0.375 | 0.750 |
| | F1 | 0.813 | 0.865 | 0.435 | 0.560 | 0.455 | 0.706 | 0.615 | 0.800 | 0.500 | 0.774 |
| | Balanced Accuracy | 0.813 | 0.844 | 0.625 | 0.688 | 0.646 | 0.814 | 0.734 | 0.859 | 0.675 | 0.857 |
| Model 5 | Accuracy | 0.781 | 0.781 | 0.833 | 0.854 | 0.846 | 0.908 | 0.838 | 0.888 | 0.908 | 0.929 |
| | Kappa | 0.563 | 0.563 | 0.571 | 0.656 | 0.526 | 0.740 | 0.356 | 0.640 | 0.592 | 0.732 |
| | Mcnemar P Value | 0.023 | 0.023 | 0.013 | 0.450 | 0.114 | 0.683 | 0.027 | 1.000 | 0.046 | 1.000 |
| | Sensitivity | 1.000 | 1.000 | 0.500 | 0.688 | 0.500 | 0.750 | 0.313 | 0.688 | 0.500 | 0.750 |
| | Specificity | 0.563 | 0.563 | 1.000 | 0.938 | 0.959 | 0.959 | 0.969 | 0.938 | 0.988 | 0.963 |
| | Precision | 0.696 | 0.696 | 1.000 | 0.846 | 0.800 | 0.857 | 0.714 | 0.733 | 0.889 | 0.800 |
| | Recall | 1.000 | 1.000 | 0.500 | 0.688 | 0.500 | 0.750 | 0.313 | 0.688 | 0.500 | 0.750 |
| | F1 | 0.821 | 0.821 | 0.667 | 0.759 | 0.615 | 0.800 | 0.435 | 0.710 | 0.640 | 0.774 |
| | Balanced Accuracy | 0.781 | 0.781 | 0.750 | 0.813 | 0.730 | 0.855 | 0.641 | 0.813 | 0.744 | 0.857 |

**Table 4** Average accuracy estimates of the RF model tested on real world data collected from Calabria and Beirut.

| Estimates | Precision | Recall | F1-score | Support |
|---|---|---|---|---|
| Class 1 | 1.00 | 0.97 | 0.98 | 37.50 |
| Class 2 | 0.98 | 1.00 | 0.99 | 67.50 |
| Accuracy | 0.99 | 0.99 | 0.99 | 0.99 |
| Macro average | 0.99 | 0.98 | 0.99 | 105.00 |
| Weighted average | 0.99 | 0.99 | 0.99 | 105.00 |